\pdfoutput=1

\documentclass[11pt]{article}

\usepackage[]{acl}

\usepackage{times}
\usepackage{latexsym}

\usepackage[T1]{fontenc}

\usepackage[utf8]{inputenc}

\usepackage{microtype}

\usepackage{booktabs}
\usepackage{arydshln}
\usepackage{graphicx}
\usepackage{caption}
\usepackage{arabtex}
\usepackage{multirow}

\newcommand{\hide}[1]{}

\newcommand{\AHAMZADN}{{\v{A}}}

\newcommand{\SHIN}{{\v{s}}}

\title{Morphosyntactic Tagging with Pre-trained Language Models\\ for Arabic and its Dialects}

\author{Go Inoue, Salam Khalifa$^\dagger$, and Nizar Habash\\
  Computational Approaches to Modeling Language (CAMeL) Lab \\
  New York University Abu Dhabi \\
  $^\dagger$Stony Brook University\\
  {\tt \{go.inoue,nizar.habash\}@nyu.edu}\\
  {\tt salam.khalifa@stonybrook.edu}
}
 
\begin{document}
\setarab
\novocalize
\maketitle
\begin{abstract}
We present state-of-the-art results on morphosyntactic tagging across different varieties of Arabic using fine-tuned pre-trained transformer language models.
Our models consistently outperform existing systems in Modern Standard Arabic and all the Arabic dialects we study, achieving 2.6\% absolute improvement over the previous state-of-the-art in Modern Standard Arabic, 2.8\% in Gulf, 1.6\% in Egyptian, and 8.3\% in Levantine.
We explore different training setups for fine-tuning pre-trained transformer language models, including training data size, the use of external linguistic resources, and the use of annotated data from other dialects in a low-resource scenario.
Our results show that strategic fine-tuning using datasets from other high-resource dialects is beneficial for a low-resource dialect. 
Additionally, we show that high-quality morphological analyzers as external linguistic resources are beneficial especially in low-resource settings.
\end{list}
\end{abstract}

\section{Introduction}
Fine-tuning pre-trained language models like BERT~\cite{devlin2019bert} has achieved great success in a wide variety of natural language processing (NLP) tasks, e.g., sentiment analysis~\cite{abu-farha-etal-2021-overview}, question answering~\cite{antoun-etal-2020-arabert}, named entity recognition~\cite{ghaddar2022jaber}, and dialect identification~\cite{abdelali2021pretraining}. 
Pre-trained LMs have also been used for enabling technologies such as part-of-speech (POS) tagging \cite{Lan:2020:empirical,khalifa-etal-2021-self,inoue:2021:interplay} to produce features for downstream processes.
Previous POS tagging results using pre-trained LMs focused on core POS tagsets;
however, it is still not clear how these models perform on the full morphosyntactic tagging task of very morphologically rich languages, where the size of the full tagset can be in the thousands.
One such language is Arabic, where lemmas inflect to a large number of forms through different combinations of morphological features and cliticization.
Additionally, Arabic orthography omits the vast majority of its optional diacritical marks which increases morphosyntactic ambiguity.


A third challenge for Arabic is its numerous variants.
Modern Standard Arabic (MSA) is the primarily written variety used in formal settings.
Dialectal Arabic (DA), by contrast, is the primarily spoken unstandardized variant.
MSA and different DAs, e.g., Gulf (GLF), Egyptian (EGY), and Levantine (LEV), vary in terms of their grammar and lexicon to the point of impeding system usability cross-dialectally \cite{Habash:2012:morphological}.
Furthermore, these variants currently differ in the degree of data availability: MSA is the highest resourced variant, followed by GLF and EGY, and then LEV.

In this paper, we explore different training setups for fine-tuning Arabic pre-trained language models in the complex morphosyntactic tagging task 
for four Arabic variants (MSA, GLF, EGY, and LEV) under controlled experimental settings.

We aim to answer the following questions:
\begin{itemize}
    \item How does the size of the fine-tuning data affect the performance?
    \item What kind of tagset scheme is suitable for modeling morphosyntactic features?
    \item Is there any additional value of using external linguistic resources?
    \item How can we make use of annotated data in some  dialects to improve performance in another low-resourced dialect?
\end{itemize}

Our system\footnote{We make our models and data publicly available at \url{https://github.com/CAMeL-Lab/CAMeLBERT_morphosyntactic_tagger}.}
achieves state-of-the-art (SOTA) performance in full morphosyntactic tagging accuracy in all the variants we study, resulting in 2.6\% absolute improvement over previous SOTA in MSA, 2.8\% in GLF, 1.6\% in EGY, and 8.3\% in LEV.
\begin{figure*}[t!]
\centering
\includegraphics[width=\textwidth]{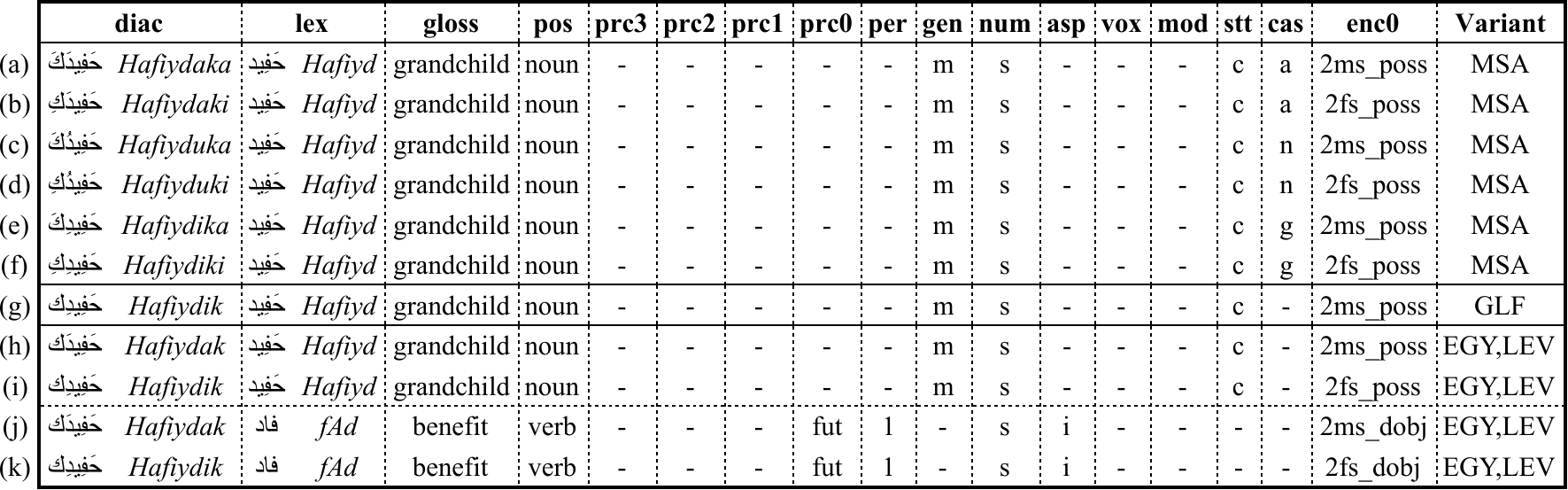}
\captionof{table}{
This is an example of multiple readings of the word <.hfydk> \emph{Hfydk} in the different variants of Arabic. The table also shows the full range of morphological features: part-of-speech (\textbf{pos}), aspect (\textbf{asp}), mood (\textbf{mod}), voice (\textbf{vox}), person (\textbf{per}), gender (\textbf{gen}), number (\textbf{num}), case (\textbf{cas}), state (\textbf{stt}) and clitics:  proclitics (\textbf{prc3, prc2, prc1, prc0}) and enclitic (\textbf{enc0}). In addition to the lemma (\textbf{lex}), fully diacritized form (\textbf{diac}), and English gloss (\textbf{gloss}).
} 
\label{tab:example}
\end{figure*}

\section{Arabic Language and Resources}

\subsection{Arabic and its Dialects}
MSA is the primarily written form of Arabic used in official media communications, official documents, news, and education. 
In contrast, the primarily spoken varieties of Arabic are its dialects.
Arabic dialects vary among themselves and can be categorized at different levels of regional classifications \cite{Salameh:2018:fine-grained}.
They are also different from MSA in most linguistic aspects (namely phonology, morphology, and syntax).
Moreover, dialects have no official status despite being widely used in different means of daily communication --  spoken as well as increasingly written on social media.
In this work, we focus on MSA, Gulf Arabic (GLF), Egyptian Arabic (EGY), and Levantine Arabic (LEV). 

\subsection{Orthography} 
In this paper, we focus on Arabic written in Arabic script for MSA and DA.
\vocalize
An important feature of Arabic orthography is the omission of diacritical marks which are mostly used to indicate short vowels and consonantal doubling.
This omission introduces ambiguity to the text, e.g., the word <ktb>~\emph{ktb}\footnote{Arabic transliteration is presented in the HSB scheme \cite{Habash:2007:arabic-transliteration}.}
could mean `to write' (<katab>~\emph{katab}) or `books' (<kutub>~\emph{kutub}) among other readings.
\novocalize

Unlike MSA, Arabic dialects have no official standard orthography.
Depending on the writer, words are sometimes spelled phonetically or closer to an MSA spelling through cognates or a mix of both.
It has been found that in extreme cases a word can have more than 20 different spellings \cite{Habash:2018:unified}.
This results in highly inconsistent and sparse datasets and models.
The Conventional Orthography for Dialectal Arabic (CODA) \cite{Habash:2018:unified} has been proposed and used in manual annotations of many datasets including some of those used in this paper.
Ideally, the process of morphological disambiguation should take raw text as input, as this is more authentic than conventionalized spelling.
We follow this principle for EGY and LEV where analyses are paired with the raw text.
However, the GLF dataset analyses are linked to the CODA version only, since orthographic conventionalization was applied as an independent step during manual data annotations and there are no simple direct mappings between the raw text and the analyses \cite{Khalifa:2018:morphologically}.

\subsection{Morphology} Arabic is a morphologically rich language where a single lemma inflects to a large number of forms through different combinations of morphological features (gender, number, person, case, state, mood, voice, aspect) and cliticization (prepositions, conjunctions, determiners, pronominal objects, and possessives).
As some of the morphological features are primarily expressed with optional diacritical marks, orthographic ambiguity results in different morphological analyses, e.g., MSA can have up to 12 analyses per word (out-of-context) on average \cite{Pasha:2014:madamira}.
MSA and DA differ in the degree of morphological complexity, for example, MSA retains nominal case and verbal mood features; but these are absent in DA.
On the other hand, many dialects take more clitics than MSA,
e.g., the <^s>+~+<mA> {\it mA+~+{\SHIN}} negation circumclitic structure found in EGY and not MSA \cite{Habash:2012:morphological}.

\vocalize
Table~\ref{tab:example} shows different possible readings for the word <.hfydk> \emph{Hfydk} among MSA, EGY, GLF, and LEV.
Rows (a) to (i) are different inflections for case or possessive pronouns or both of the lemma <.hafiyd> \emph{Hafiyd} `grandchild' for all variants.
Rows (j) and (k) show different readings that are inflections of the verb lemma <fAd> \emph{fAd} `to benefit', the inflections are for different object pronouns.
Note that even between the different POS inflections words can sound and look exactly the same, this shows the degree of morphological complexity and ambiguity in Arabic and its dialects.

\subsection{Resources}
\label{subsec:resources}

\begin{table}[t!]
\centering
\includegraphics[scale=0.69]{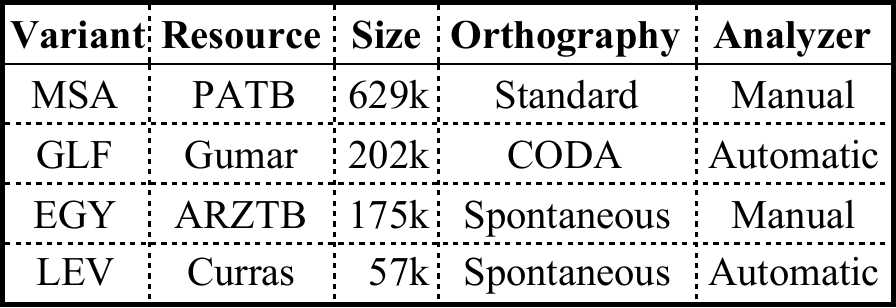}
\caption{
An overview of the current status of the data and morphological analyzers used in this work.
} 
\label{tab:overview}
\end{table}

In this work, we use datasets that have been fully annotated for morphological features and cliticization among other lexical features such as lemmas.
We use the Penn Arabic Treebank for MSA~\cite{Maamouri:2004:patb}, ARZTB~\cite{Maamouri:2012:arz} for EGY, the Gumar corpus~\cite{Khalifa:2018:morphologically} for GLF, and the Curras corpus~\cite{Jarrar:2014:building} for LEV.
We also use morphological analyzers that provide out-of-context analyses for a given word, those analyzers provide the same set of features that are seen in the annotated data.
For MSA we use the SAMA database \cite{Graff:2009:standard}, and for EGY we use CALIMA \cite{Habash:2012:morphological}.
Both GLF and LEV do not have morphological analyzers, instead, we use automatically generated analyzers from their training data using paradigm completion as described in \citet{Eskander:2013:automatic-extraction,Eskander:2016:creating} and \citet{khalifa-etal-2020-morphological}.
The quality and coverage of analyzers, in general, can differ depending on how they were created.
Manually created analyzers (MSA and EGY in this work) tend to have a better quality and lexical coverage over automatically created ones (GLF and LEV in this work).
The quality of automatically generated analyzers is also highly dependent on the quality and size of the training data used to create them.

Table~\ref{tab:overview} shows the overall state of the resources for each dialect studied in this work.
In terms of the size of fully annotated corpora in tokens, MSA is approximately three times larger than GLF and EGY and 11 times larger than LEV.
Both MSA and GLF have consistent orthography whereas EGY and LEV are more noisy.
When it comes to external morphological analyzers, only MSA and EGY have manually created and checked morphological analyzers, while both GLF and LEV have analyzers created automatically.
This contrast of resource availability allows us to study how challenging the morphosyntactic tagging task can be in different real-world situations.
\section{Related Work}
Arabic morphological modeling proved to be useful in a number of downstream NLP tasks such as machine translation  \cite{Sadat:2006:combination,ElKholy:2012:orthographic}
speech synthesis \cite{halabi2016modern}, dependency parsing \cite{Marton:2013:dependency}, sentiment analysis \cite{Baly:2017:sentiment}, and gender reinflection \cite{alhafni-etal-2020-gender}.
We expect all of these applications and others to benefit from improvements in morphosyntactic tagging.

There have been multiple approaches to morphological modeling for Arabic.
Those approaches differ depending on the target tagset (POS vs full morphology) and the availability of linguistic resources.
When it comes to MSA and DA full morphological tagging, MADAMIRA \cite{Pasha:2014:madamira} trained separate SVM taggers for each morphological feature (including cliticization) and selected the most probable answer provided by an external morphological analyzer all in one step for both MSA and EGY.
AMIRA \cite{Diab:2004:automatic} on the other hand used a cascading approach where it performed POS tagging after automatically segmenting the text.

A more recent similar approach to MADAMIRA was introduced by \newcite{Zalmout:2017:dont} but using a neural architecture instead.
\newcite{Inoue:2017:joint} presented a multitask neural architecture that jointly models individual morphological features for MSA.
\newcite{zalmout-habash-2019-adversarial} extended \newcite{Zalmout:2017:dont}'s work using multitask learning and adversarial training for full morphological tagging in MSA and EGY.
Similarly, \newcite{zalmout-habash-2020-joint} proposed an approach where they jointly model lemmas, diacritized forms, and morphosyntactic features, providing the current state-of-the-art in MSA.
The same approach was used in \newcite{khalifa-etal-2020-morphological}, where they focused on the effect of the size of the data and the available linguistic resources and the impact on the overall performance on morphosyntactic tagging for GLF.
\newcite{zalmout2020} provides the current state-of-the-art performance in LEV by extending \newcite{khalifa-etal-2020-morphological}'s work to LEV.

Another line of research that works with DA includes \newcite{Darwish:2018:multi-dialect}, where they presented a multi-dialectal CRF POS tagger, using a small set of 350 manually annotated tweets for each of EGY, GLF, LEV, and Maghrebi Arabic \cite{samih-etal-2017-learning}.
We do not evaluate on their data because their task is defined as shallow morpheme segmentation and tagging; this is quite different from, and not easily mappable to, our task, where we disambiguate morphosyntactic features of the whole word without identifying its morpheme segments.
Additionally, their tagset includes social media specific tags, such as HASH, EMOT, and MENTION, which are not in any of the large standard dataset and analyzers we study in this paper.

Pre-trained LM-based efforts in Arabic morphosyntactic tagging are relatively limited and either assume gold segmentation or only produce core POS tags.
\newcite{kondratyuk-2019-cross} leveraged the multilingual BERT model with additional word-level and character-level LSTM layers for lemmatization and morphological tagging, assuming gold segmentation.
They reported the results for the SIGMORPHON 2019 Shared Task~\cite{mccarthy-etal-2019-sigmorphon}, which includes MSA.
\newcite{inoue:2021:interplay} reported POS tagging results in MSA, GLF, and EGY using BERT models pre-trained on Arabic text with various pre-training configurations.
They do not assume pre-segmentation of the text, however, they only consider the core POS tag, rather than the fully specified morphosyntactic tag.
\newcite{khalifa-etal-2021-self} proposed a self-training approach for core POS tagging where they iteratively improve the model by incorporating the predicted examples into the training set used for fine-tuning.

In this paper, we work with full morphosyntactic modeling on unsegmented text in four different variants of Arabic: MSA, GLF, EGY, and LEV.
Furthermore, we explore the behavior of the pre-trained LM with respect to fine-tuning data size under different training setups.
Given the available resources, we recognize our results' limitations in terms of applicability to different genres and styles, as well as noisy social media text and Roman script Arabic text \cite{Darwish:2014:arabizi}.

\section{Methodology}

\subsection{Morphosyntactic Tagging with Pre-trained LMs}
To obtain a fully specified morphosyntactic tag sequence, we build a classifier for each morphosyntactic feature independently, inspired by MADAMIRA.
Unlike MADAMIRA where they use an SVM classifier, we use two pre-trained LM based classifiers: CAMeLBERT-Mix for DA and CAMeLBERT-MSA for MSA~\cite{inoue:2021:interplay}.
In selecting these pre-trained language models, we considered the results from \newcite{inoue:2021:interplay} who showed that CAMeLBERT-Mix, their largest Arabic BERT model by training data size,  gives the best results on DA tasks.  
CAMeLBERT-MSA, which  outperforms  CAMeLBERT-Mix on MSA tasks, is only second to AraBERT~\cite{antoun-etal-2020-arabert}, but since it was created under the same setting as CAMeLBERT-Mix, it minimizes experimental variations in our study.\footnote{
We leave engineering optimization using other pre-trained language models to future work.}
Following the work of \newcite{devlin2019bert}, fine-tuning the CAMeLBERT models is done by appending a linear layer on top of its architecture.
We use the representation of the first sub-token as an input to the linear layer.

\subsection{Factored and Unfactored Tagset}
One of the challenges of the morphosyntactic tagging is the large size of the full tagset due to morphological complexity of the language, where a complete single tag is a concatenation of all the morphosyntactic features.
For example, MSA and EGY data have approximately 2,000 unique complete tags in the training data, whereas GLF and LEV have around 1,400 and 1,000 tags, respectively. These are not the full tagsets as there are many feature combinations that are not seen in the data.

MADAMIRA's basic approach is to use a factored feature tagset that comprises multiple tags, each representing a corresponding morphosyntactic category.\footnote{
For example, the tagset for MSA comprises POS (34 tags), per (4), gen (3), num (5), asp (4), vox (4), mod (5), stt (5), cas (5), prc3 (3), prc2 (9), prc1 (17), prc0 (7), enc0 (48).
}
This approach remedies the issue of the large tagset size by dividing it into multiple sub-tagsets of small sizes, however, it may produce inconsistent tag combinations.

Alternatively, one can combine the individual tags into a single tag.
This approach has the advantage of guaranteeing the consistency of morphosyntactic feature combinations.
However, it may not be optimal in terms of tag coverage due to a large number of unseen tags in the test data in addition to the large space of classes.

To determine which approach is most suitable for modeling, we build morphosyntactic taggers with both the factored tagset and the unfactored tagset for each variant.
Additionally, we explore the effect of the training data size for both settings.

\subsection{Retagging via Morphological Analyzers}
\label{subsec:retagging}
In previous efforts~\cite{Zalmout:2017:dont,khalifa-etal-2020-morphological}, it has been shown that lexical resources such as morphological analyzers can boost the performance of morphosyntactic tagging through an in-context ranking of out-of-context answers provided by the analyzer.

In this work, we follow their approach, where we use the morphological analyzers as a later step after tagging with the fine-tuned pre-trained model.
We use the analyzers described in Section~\ref{subsec:resources} to provide out-of-context analyses.
For each word, the analyzer may provide more than one answer.\footnote{Both the MSA and EGY analyzers provide backoff modes.
We use the recommended setting by \newcite{Zalmout:2017:dont}.
For GLF and LEV analyzers we keep the original predictions if no answer is returned.}
The analyses are then ranked based on the unweighted sum of successful matches between the values of the predictions from the individual taggers and those provided by the analyzer.
To break ties during the ranking, we take the weighted sum of the probability of the \emph{unfactored} feature tag and the product of the probabilities of all the individual tags as follows:

\begin{equation}
\frac{1}{2}P(t_{unfactored})+ \frac{1}{2}\prod_{m \in M}P(t_{m})
\label{eq:tie1}
\end{equation}
where $t$ is the tag for the feature $m$ and $M$ is the set of morphosyntactic features.
The probabilities are obtained through unigram models based on the respective training data split.

\subsection{Merged and Continued Training}
Morphosyntactic modeling for DA is especially challenging because of  data scarcity.
Among the datasets that we use, LEV is the least resourced variant, having 11 times less training data than MSA.
Therefore, we want to investigate an optimal approach to utilize data from other variants to improve upon the performance of morphosyntactic tagging for LEV.

In this work, we experiment with the following two settings:
(a) we merge all the datasets together and fine-tune a pre-trained LM on the merged datasets in a single step; and 
(b) similar to \newcite{zalmout2020}, we start fine-tuning a pre-trained LM on a mix of high-resource datasets (MSA, GLF, and EGY), and then continue fine-tuning on a low-resource dataset (LEV).
\section{Experiments}
\label{sec:experiments}
\subsection{Experimental Settings}
\label{subsec:experimental_settings}
\paragraph{Data}
To be able to compare with previous SOTA \cite{zalmout-habash-2020-joint,zalmout-habash-2019-adversarial,khalifa-etal-2020-morphological,zalmout2020}, we follow the same conventions they used for data splits: MSA and EGY \cite{Diab:2013:ldc}, GLF \cite{Khalifa:2018:morphologically}, and LEV \cite{Eskander:2016:creating}.
In Table \ref{tab:data_split}, we show the statistics of our datasets.

\begin{table}[t!]
\centering
\includegraphics[width=0.33\textwidth]{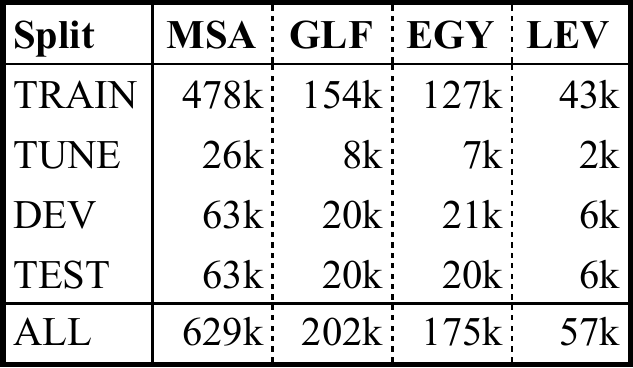}
\caption{
Statistics on TRAIN, TUNE, DEV, and TEST for each variant in terms of number of words.
} 
\label{tab:data_split}
\end{table}

\begin{table*}[t!]
\centering
\includegraphics[scale=0.69]{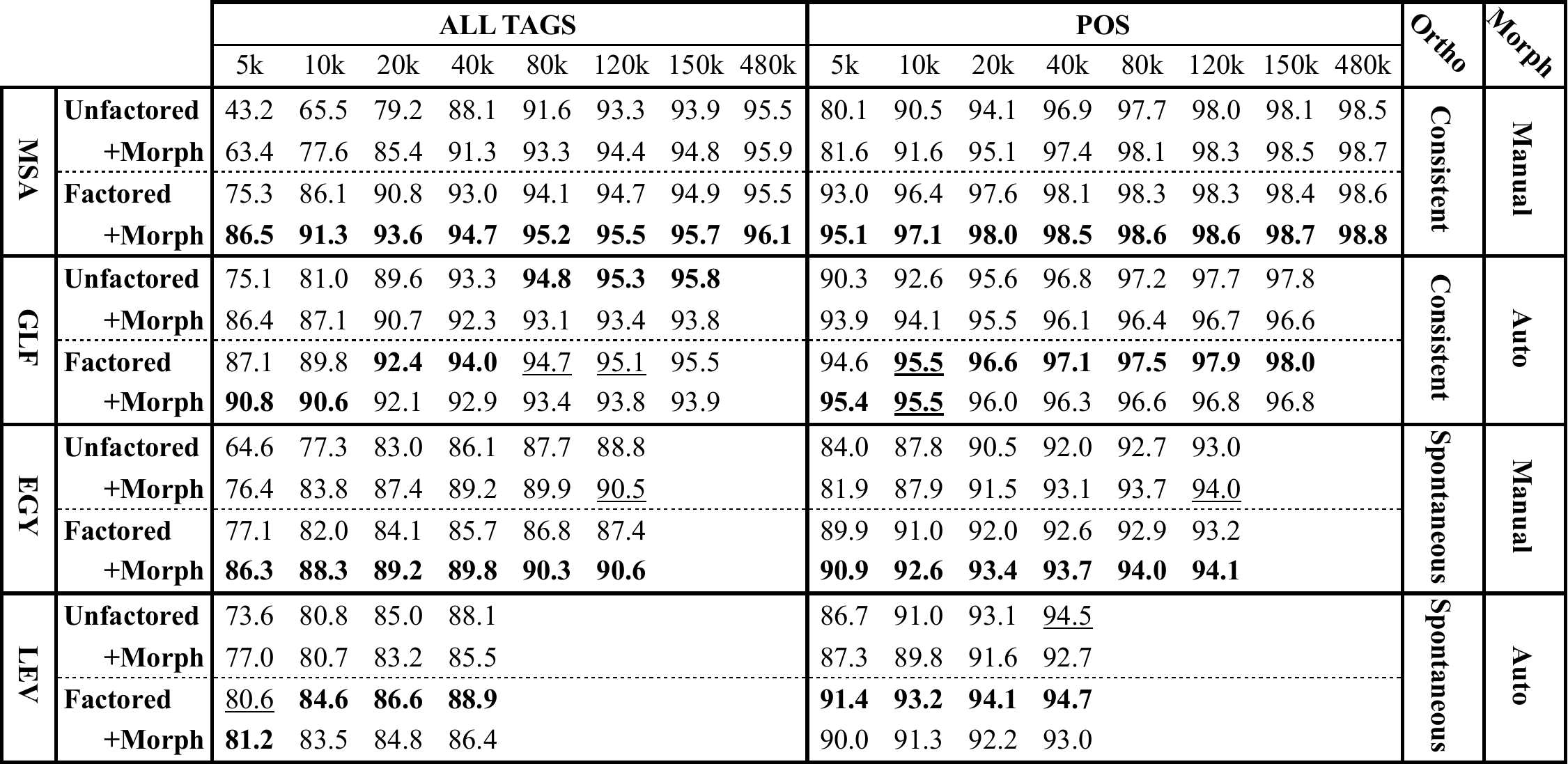}
\caption{
\textsc{DEV} results on a learning curve of the training data size.
Morph refers to the model with an additional step of retagging using a morphological analyzer.
We bold the best score for each variant.
Underlined scores denote that the differences between those scores and the best scores are statistically insignificant with McNemar’s test ($p< 0.05$).
}
\label{tab:table4_main}
\end{table*}

\paragraph{Fine-tuning}
We fine-tuned the CAMeLBERT models \cite{inoue:2021:interplay} on each morphosyntactic tagging task.
Following their recommendation, we used CAMeLBERT-MSA for MSA and CAMeLBERT-Mix for the dialects.
We used Hugging~Face's transformers~\cite{wolf2020huggingfaces} for implementation.
We trained our models for 10 epochs with a learning rate of 5e-5, a batch size of 32, and a maximum sequence length of 512.
We pick the best checkpoint based on TUNE and report results on DEV and TEST from a single run.

\paragraph{Learning Curve}
To investigate the effect of fine-tuning data sizes, we randomly sample training examples on a scale of 5k, 10k, 20k, 40k, 80k, 120k, and 150k tokens.
We use 150k, 120k, and 40k since they are comparable to the number of tokens in GLF, EGY, and LEV datasets, respectively.
This allows us to measure the performance difference across different dialects in a controlled manner.
This also gives us insight into the amount of annotated data required to achieve a certain performance, which is useful when creating annotated resources for new dialects.
We use this setup in all the reported experiments.

\paragraph{Pre-processing for Merged and Continued Training}
Although the different datasets provide the same set of morphosyntactic features, there exist some inconsistencies between them.
The datasets were annotated by different groups using slightly different annotation guidelines, therefore, we need to bring all the feature values into a common space with LEV.
We performed the following steps to address those inconsistencies:
(a) we drop the \textit{stt, cas, mod, vox, enc1}, and \textit{enc2} features;
(b) we remove the diactization from the lexical parts of the proclitic features, e.g., the conjunction +\<w> {\it w+} realized as \emph{wa\_conj} in MSA and \emph{wi\_conj} in EGY both maps to \emph{w\_conj} in LEV; and
(c) for certain POS classes some features have default values in case they are not present, those default values were different for different datasets. Thus, we mapped those default values to match whatever was specified as default in LEV.
We only performed these modifications for the experiments on merged and continued training.

\paragraph{Evaluation Metrics}
We compute the accuracy in terms of the core POS and the combined morphosyntactic features (\textbf{\textsc{ALL TAGS}}).
For MSA, we use 14 features, which are \textit{pos, per, gen, num, asp, vox, mod, stt, cas, prc3, prc2, prc1, prc0}, and \textit{enc0}.
For dialects, we use 16 features, where we include \textit{enc1} and \textit{enc2} in addition to the 14 features used in MSA.
In the merged and continued training setup, we use a reduced set of
10 features, \textit{pos, per, gen, num, asp, prc3, prc2, prc1, prc0}, and \textit{enc0}, which we refer to as \textbf{\textsc{ALL TAGS 10}}.

\subsection{Results}
\paragraph{Factored vs Unfactored Models}
Table \ref{tab:table4_main} shows the DEV results for the models trained with the factored and unfactored tagset (henceforth, factored and unfactored models, respectively) on a learning curve of the training data size.
In the extremely low-resource setting of 5k tokens in the \textsc{ALL TAGS} metric, we observe that factored models consistently outperform unfactored models across all the variants (15.9\% absolute increase on average).
In particular, MSA benefited most with a 32.1\% absolute increase, followed by EGY (12.5\%), GLF (12.0\%), and LEV (7.1\%).

However, this gap shrinks as the data size increases.
For instance in MSA, the differences between the scores of the factored model and the unfactored model become statistically insignificant by McNemar's test~\cite{McNemar1947} with $p<0.05$ when trained on the full data.
This is presumably due to the decrease in the number of unseen unfactored tags in DEV.
In fact, 3.9\% of the unfactored tags in DEV are not seen in TRAIN in the 5k setting, whereas only 0.1\% of tags are unseen in DEV when we use the full data.

The factored model performs better than the unfactored model across all the data sizes in MSA and LEV.
The EGY and GLF models follow a similar pattern in the low resourced settings, however, the unfactored models begin to perform better than the factored ones from 20k for EGY and 40k for GLF.
Our results suggest that the factored tagset is optimal compared to the unfactored tagset, especially in low-resource settings.

\paragraph{Retagging with Morphological Analyzer}
We observe that the use of a morphological analyzer consistently improves the performance of both unfactored and factored models across all the different training data sizes in MSA and EGY in \textsc{ALL TAGS}.
The value of a morphological analyzer is especially apparent in the very low resourced setting (5k), with an increase of 20.2\% (MSA) and 11.8\% (EGY) in the unfactored model and 11.2\% (MSA) and 9.2\% (EGY) in the factored model.
However, the effect of retagging with a morphological analyzer diminishes as the data size increases, yet providing a performance gain of 0.4\% in the unfactored model with the analyzer and 0.5\% in its factored counterpart in the high resourced setting in MSA.

Similarly, we observe an increase in performance when we include a morphological analyzer in the very low-resourced settings in GLF and LEV.
However, as we increase the training data size, the use of a morphological analyzer starts to hurt the performance at 40k in GLF and 10k in LEV in the unfactored model and 20k in GLF and 10k in LEV in the factored model.
We observe here that the quality of the analyzer has direct implications on the performance. The analyzers used for MSA and EGY are of higher quality since they were manually created and checked, whereas GLF and LEV analyzers are impacted by the quality and size of the annotated data used to create them. This is also consistent with the findings of \newcite{khalifa-etal-2020-morphological}.

\begin{table*}[t!]
\centering
\includegraphics[scale=0.69]{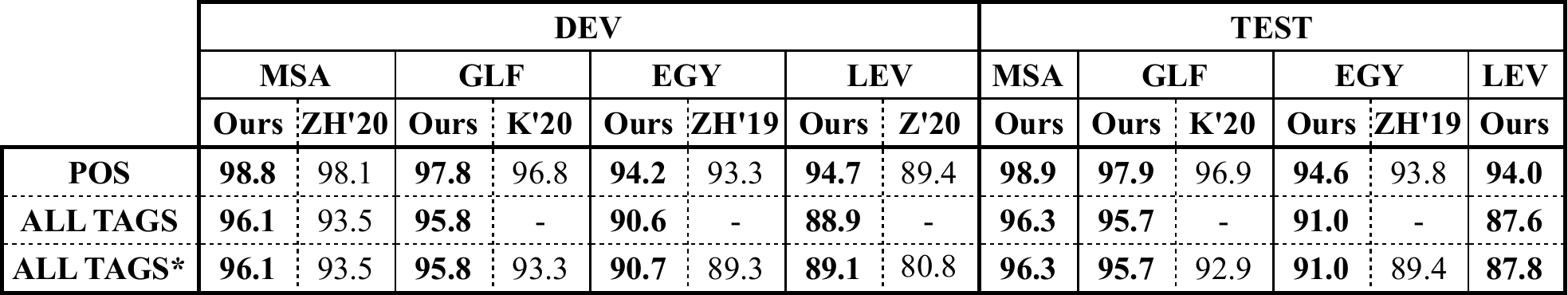}
\caption{
DEV and TEST results of our systems and previously published systems on the same datasets.
} 
\label{tab:table5_comparison}
\end{table*}

\paragraph{Comparison with Previous SOTA Systems}
Table \ref{tab:table5_comparison} shows DEV and TEST results for our models and a number of previously published state-of-the-art morphosyntactic tagging systems.
For our models, we use the best systems in terms of \textsc{ALL TAGS} metric, namely, the factored model with a morphological analyzer for MSA and EGY, the unfactored model for GLF, and the factored model for LEV.
For existing models, we report the best results from \newcite{zalmout-habash-2020-joint} (ZH'20) for MSA, \newcite{khalifa-etal-2020-morphological} (K'20) for GLF, \newcite{zalmout-habash-2019-adversarial} (ZH'19) for EGY,
and \newcite{zalmout2020} (Z'20) for LEV.

Since some of these systems do not report on all of the features that we report on, but rather on different subsets of them, we include in the table our results when matched with their features (\textsc{ALL TAGS*} in Table~\ref{tab:table5_comparison}).
There is no difference for MSA; however the \textsc{ALL~TAGS*} setting for EGY and LEV excludes \textit{enc1} and \textit{enc2}. As for GLF,  \textsc{ALL~TAGS*} consists of only 10 features: \textit{pos, asp, per, gen, num, prc0, prc1, prc2, prc3,} and \textit{enc0}.

We observe that our models consistently outperform the existing systems in all variants.
Our model achieves 2.6\% absolute improvement over the state-of-the-art system in MSA, 2.8\% in GLF, 1.6\% in EGY, and 8.3\% in LEV.

\begin{table*}[t!]
\centering
\includegraphics[scale=0.69]{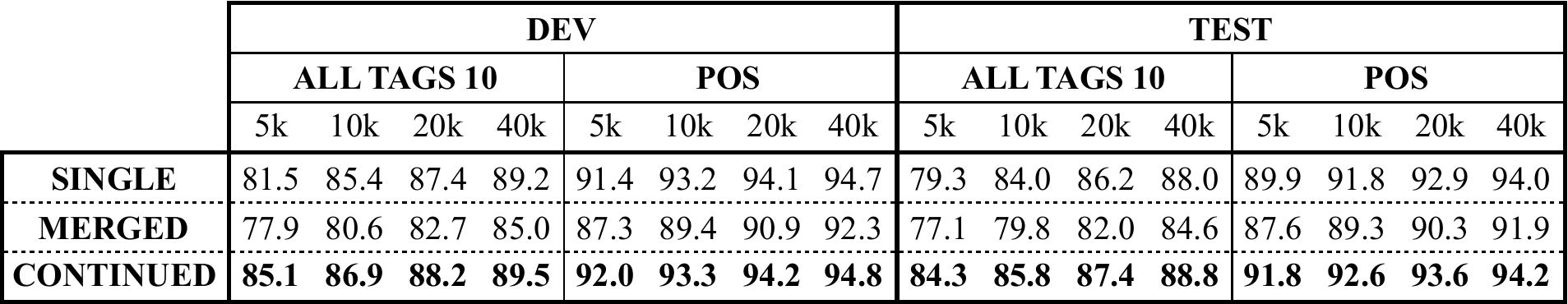}
\caption{
\textsc{DEV} and \textsc{TEST} results on LEV for the merged training setup (MERGED) and the continued training setup (CONTINUED).
SINGLE refers to the model trained only on LEV.
} 
\label{tab:table6_strategy}
\end{table*}

\paragraph{Merged and Continued Training}
Table \ref{tab:table6_strategy} shows the results on LEV for the merged and the continued training setups.
We use the factored model without the analyzer as it was the best setup in the experiments presented so far.
The results for merged training are consistently below those for the baseline across different data sizes, even though they have access to more data.
This is most likely a result of the disproportionately small size of the LEV dataset when compared to the other variants.

In contrast, the results for continued training show consistent improvements over the LEV-only baseline model.
Continued training provides a substantial increase in performance, especially in the very low resourced setting with only 5k tokens, giving 3.6\% absolute improvement over the baseline on the DEV set.
Our results show that continued training from the model trained on high-resourced dialects is very beneficial with lower amounts of training data.
These results are not directly comparable to the previous SOTA because of the different training data and metric used.

\subsection{Error Analysis}
\paragraph{OOV}
To better understand the effect of different training setups, we examine the performance of our models on out-of-vocabulary (OOV) tokens alone.
Here, we observe a stronger and more consistent pattern.
The \textit{average} difference between the best model and the weakest model in ALL TAGS across variants is larger in OOV tokens (6.7\%) than in all tokens (2.3\%).
On OOV tokens, the factored model with a morphological analyzer consistently performs best in all the data sizes for all the variants except for LEV.
In LEV, however, the same model without the morphological analyzer outperforms the one with the analyzer.
This is presumably due to the orthographic inconsistency in the data along with the quality of the morphological analyzer as discussed in Section~\ref{subsec:resources}.

\begin{table*}[ht!]
\centering
\includegraphics[scale=0.69]{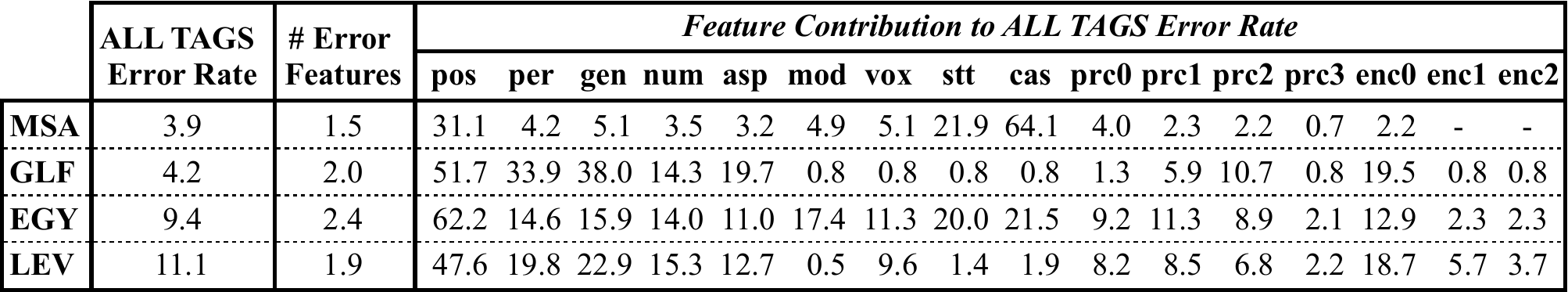}
\caption{
The number and percentage of specific feature errors among the ALL TAGS errors in the best systems on the DEV set.
} 
\label{tab:table_error_analysis}
\end{table*}

\paragraph{Error Statistics}
Table \ref{tab:table_error_analysis} presents the number and percentage of specific feature errors among the ALL TAGS errors in the best systems on the DEV set.
On average, there are two feature prediction failures within an unfactored tag across the different variants.
We observe that MSA and DA exhibit different error patterns:
In MSA, case is the largest error contributor among other features, which is consistent with the previous findings along the line~\cite{zalmout-habash-2020-joint}, whereas in dialects, POS is the largest error contributor.

Among the POS errors, the most common error type is mislabeling a nominal tag with a different nominal tag, at 44.2\% of the errors in GLF, 67.3\% in EGY, and 57.8\% in LEV, while this type of error is more dominant in MSA (80.8\%).
Mislabeling nominals with verbs is more common in DA at 23.1\% in GLF, 13.0\% in EGY, and 20.1\% in LEV, compared to MSA (7.7\%).

The core morphological features such as \textit{per, gen, num}, and \textit{asp} have a higher percentage of errors in DA than in MSA.
Another noticeable difference is \textit{enc0} feature (MSA $\sim$2\% vs DA on average $\sim$17\%).
This is likely due to label distribution differences in pronominal enclitics: MSA has a highly skewed distribution with 90\%, 1\%, and 9\% ratio for 3rd, 2nd and 1st persons as expected in MSA news genre.
In comparison, DA has less skew with 50\%, 17\%, and 32\% respectively, which increases the likelihood of error. 

Among the three dialects, we observe similar patterns in terms of feature error contribution, especially for GLF and LEV with a correlation coefficient of 0.93.
However, in EGY specifically, we observe a high percentage of errors in \textit{mod, vox, stt}, and \textit{cas}, partly due to the difference and inconsistency in annotation schemes.

\novocalize
We also found some gold errors which affect all of the systems we compared (previous SOTA and ours).
For example, there are cases where genitive diptotes are annotated as accusative,\footnote{For more information on Arabic morphology in a computational context, see \newcite{Habash:2010:introduction}.} e.g., the word <'iyrAn>~\emph{{\AHAMZADN}yrAn} `Iran' in the context <fy 'iyrAn> ~\emph{fy {\AHAMZADN}yrAn} `in Iran'.
As the results on Arabic morphosyntactic disambiguation are reaching new heights, it may be useful for the community using these resources to revisit their annotations.

\section{Conclusion and Future Work}
In this paper, we presented the state-of-the-art results in the morphosyntactic tagging task for Modern Standard Arabic and three Arabic dialects that differ in terms of linguistic properties and resource availability.
We conducted different experiments to examine the performance of pre-trained LMs under different fine-tuning setups.
We showed that the factored model outperforms the unfactored model in low-resource settings; however, this gap diminishes as the data size increases.
Additionally, high-quality morphological analyzers proved to be helpful, especially in low-resource settings.
Our results also show that fine-tuning using datasets from other dialects followed by fine-tuning using the target dialect is beneficial for low-resource settings.  
Our systems outperform previously published SOTA on this task.

In the future, we plan to investigate continued training further and find other ways where we can utilize resources and datasets for low-resourced dialects.
We also intend to explore other architectures for morphosyntactic tagging using multi-task learning in the context of pre-trained LMs, as well as work on the task of automatic lemmatization.
We also plan to integrate some of our best models as part of the Python open-source Arabic NLP toolkit CAMeL~Tools~\cite{obeid-etal-2020-camel}.

\section{Ethical Considerations}
The experiments reported in this work rely on previously published datasets described in Section~\ref{subsec:resources}.
We used the CAMeLBERT models along with morphosyntactically annotated datasets to build our morphosyntactic taggers, which is in line with their intended use.
Our work is on core and generic NLP technologies that can be potentially used with malicious intent, for example, as part of the pipeline.
To ensure reproducibility, we make our code publicly available.
The details on the datasets and training are described in Appendix~\ref{sec:appendix}.
Given the focus of this paper and the available resources, we recognize the limitations of our findings in terms of applicability to different genres, styles, and other languages.

%

\section*{Acknowledgment}
This work was carried out on the High Performance Computing resources at New York University Abu Dhabi.
We thank anonymous reviewers for their insightful suggestions and comments.
We thank Bashar Alhafni and Ossama Obeid for their assistance with the codebase and the helpful discussions.

\bibliography{anthology,camel-bib-v2,custom}
\bibliographystyle{acl_natbib}

\appendix
\section{Replicability}
\label{sec:appendix}

\subsection{Resources}
\label{appendix:resources}
\paragraph{Pretrained transfromer models}
We fine-tuned CAMeLBERT-MSA for the morphosyntactic tagging task in MSA and CAMeLBERT-Mix~\cite{inoue:2021:interplay} for EGY, GLF, and LEV.

\paragraph{Fine-tuning Data}
We used the Penn Arabic Treebank for MSA~\cite{Maamouri:2004:patb}, ARZTB~\cite{Maamouri:2012:arz} for EGY, the Gumar corpus~\cite{Khalifa:2018:morphologically} for GLF, and the Curras corpus~\cite{Jarrar:2014:building} for LEV.
The preprocessing of the data includes fixing inconsistent annotations and removing diacritics through CAMeL~Tools~\cite{obeid-etal-2020-camel}.
This preprocessing was followed in all the previous work we compared with \newcite{zalmout-habash-2019-adversarial, zalmout-habash-2020-joint, khalifa-etal-2020-morphological, zalmout2020}.

\paragraph{Data Sampling}
For the learning curve experiment in Section~\ref{subsec:experimental_settings},
we sampled the training data up to 5k, 20k, 40k, 80k, 120k, 150k tokens after shuffling the entire dataset.
Each sample after 5k is inclusive of the smaller samples.

\paragraph{Morphological Analyzers}
The morphological analyzers used in our experiments are the following:
For MSA we use the SAMA database \cite{Graff:2009:standard}, and for EGY we use CALIMA \cite{Habash:2012:morphological}.
For GLF and LEV, we use automatically generated analyzers from their training data using paradigm completion as described in \citet{Eskander:2013:automatic-extraction,Eskander:2016:creating} and \citet{khalifa-etal-2020-morphological}.

\paragraph{Data Accessibility}
MSA and EGY related resources need a license from the Linguistic Data Consortium (LDC).
The GLF data is available at \url{http://resources.camel-lab.com/} and the LEV data is available at \url{https://portal.sina.birzeit.edu/curras/}.
We provide conversion scripts to generate our preprocessed datasets from legally accessed third-party datasets at \url{https://github.com/CAMeL-Lab/CAMeLBERT_morphosyntactic_tagger}.



\subsection{Implementation}
We used Hugging Face's transformers~\cite{wolf2020huggingfaces} for implementation.
Fine-tuning is done by adding a fully connected linear layer to the last hidden state.
We release our code including the hyperparameters used in the experiments at \url{https://github.com/CAMeL-Lab/CAMeLBERT_morphosyntactic_tagger}.

For the experiments in Section \ref{sec:experiments}, we use the following hyperparameters:
a random seed of 12345, training for 10 epochs, saving the model for every 500 steps, a learning rate of 5e-5, a batch size of 32, and a maximum sequence length of 512.
We pick the best checkpoint based on TUNE and report results on DEV and TEST from a single run.

The number of parameters of the factored model for MSA is about 1.5 billion, while the factored model for GLF, EGY, and LEV has 1.8 billion parameters in total.
The unfactored model has about 110 million parameters for MSA, GLF, EGY, and LEV.

The factored model is the most computationally expensive model to train, which took about 21 hours for MSA, 16 hours for GLF, 13 hours for EGY, and five hours for LEV on a single NVIDIA-V100 card.
The unfactored model took about 90 minutes to train for MSA, 60 minutes for GLF, 50 minutes for EGY, and 20 minutes for LEV on the same machine.

\end{document}